\documentclass[conference]{IEEEtran}
\usepackage{lineno} % Add line numbers
\IEEEoverridecommandlockouts
\usepackage{cite}
\usepackage{amsmath,amssymb,amsfonts}
\usepackage{algorithmic}
\usepackage{graphicx}
\usepackage{textcomp}
\usepackage{ifthen}
\usepackage{enumitem}
\usepackage{hyperref}
\usepackage{pifont}
\usepackage{rotating}
\usepackage{ulem}  % Without the 'normalem' option
\usepackage{svg}
\usepackage{svg-extract}
\usepackage{balance}  
\usepackage{lipsum}
\usepackage{hyperref}
\usepackage{xspace}
\usepackage{enumitem}
\usepackage{float}
\usepackage{tcolorbox}
\usepackage{graphicx}    % For including images
\usepackage{subcaption}  % For subfigures
\usepackage{booktabs}  % For better horizontal lines
\usepackage{placeins} % Add this in the preamble
\usepackage[table,xcdraw]{xcolor} % Combined all xcolor options here
\usepackage{array}     % For more control over column alignment
\usepackage{setspace} % Package to control line spacing
\usepackage{wrapfig}
\hypersetup{
    colorlinks=true,
    linkcolor=blue,
    filecolor=magenta,      
    urlcolor=cyan,
    pdftitle={Overleaf Example},
    pdfpagemode=FullScreen,
    }
\usepackage{amsmath} % for the equation* environment
\usepackage{stfloats}  % Include this in your preamble
\usepackage{svg}  % Include this in the preamble to support SVG images
\usepackage{array}
\usepackage{multirow}
\usepackage{graphicx}
\tcbuselibrary{listingsutf8} % Ensures UTF-8 compatibility for listings
\usepackage{soul}
\usepackage{lscape} % To make the page landscape
\usepackage{tikz} % For drawing custom checkboxes
\usepackage{tabularx} % For tables with automatic text wrapping
\usepackage{caption}  % For better control of table captions
\usepackage{array}    % For customizing column alignment
\usepackage{multicol} % For multi-column text
\usepackage{balance}  % For balancing columns (optional)
\usepackage{float} 
\definecolor{BlueGray}{RGB}{182,205, 216}
\definecolor{DarkBlueGray}{RGB}{63, 82, 92}
\definecolor{GreenGray}{RGB}{77,107,83}
\definecolor{safagreen}{HTML}{659447}
\newcolumntype{L}[1]{>{\raggedright\let\newline\\\arraybackslash}p{#1}} 
\newcolumntype{C}[1]{>{\centering\let\newline\\\arraybackslash}p{#1}} 
\newcolumntype{R}[1]{>{\raggedleft\let\newline\\\arraybackslash}p{#1}}

\def\BibTeX{{\rm B\kern-.05em{\sc i\kern-.025em b}\kern-.08em
    T\kern-.1667em\lower.7ex\hbox{E}\kern-.125emX}}

\newcommand{\selfm}{\textit{self}}
\begin{document}

\title{A Model for Mediating Multi-Modal Human Intent into Safe Maneuvers for UAVs}

\author{\author{
    \IEEEauthorblockN{Sofia Nelson}
    \IEEEauthorblockA{
    \textit{University of Notre Dame} \\  
    Notre Dame, IN, USA \\  
    snelso24@nd.edu}  
    \and  
    \IEEEauthorblockN{Dalal Alrajeh}  
    \IEEEauthorblockA{ 
    \textit{Imperial College London} \\  
    London, UK \\  
    dalal.alrajeh@ic.ac.uk}  
    \and  
    \IEEEauthorblockN{Pedro Antonio Alarcon Granadeno}  
    \IEEEauthorblockA{
    \textit{University of Notre Dame} \\  
    Notre Dame, IN, USA \\  
    palarcon@nd.edu}  
    \and
    \IEEEauthorblockN{Jane Cleland-Huang}  
    \IEEEauthorblockA{
    \textit{University of Notre Dame} \\  
    Notre Dame, IN, USA \\  
    janehuang@nd.edu}  
}}

\maketitle

\begin{abstract}

Direct human interaction with autonomous UAV systems can be enabled through modalities such as speech, gestures, and graphical interfaces. However, interpreting such inputs as directly executable commands introduces safety risks in dynamic environments. Operator requests may conflict with terrain constraints, inter-UAV separation requirements, or flight-envelope limitations. In this paper, we present a requirements-governed maneuver-response model that mediates multi-modal human intent into safe UAV maneuvers by treating operator inputs as bounded maneuver requests rather than direct commands. Requested maneuvers are mapped to constrained motion primitives and processed through a structured request–evaluate–execute pipeline. Each request is interpreted with associated confidence, validated against terrain, separation, workspace, and flight-envelope constraints, and either constrained, rejected, or executed under continuous runtime monitoring.
We further formalize the approach as a requirements-based specification model in which maneuver primitives are associated with explicit preconditions, invariants, guard conditions, and postconditions governing admissibility, execution safety, and emergency handling. These requirements support runtime verification and future reactive synthesis approaches. We present an initial lab-based validation demonstrating that voice- and GUI-based inputs can be reliably interpreted and safely executed as constrained maneuver requests.
\end{abstract} 

\section{Motivation}
\label{sec:motivation}
Within the Human-Machine Teaming (HMT) paradigm, small Uncrewed Aerial Vehicles (UAVs) operate as collaborative partners that combine autonomous decision-making with human oversight and intervention \cite{DBLP:conf/seams/Cleland-HuangAV22}. Although these systems are often deployed under human-on-the-loop (HoTL) supervision, operators may periodically engage more directly in the loop by issuing maneuver- or mission-level commands through speech, gestures, or hardware and software interfaces. Such interventions are especially important in dynamic operational contexts where human judgment, situational awareness, or mission intent may temporarily supersede autonomous behavior. For instance, an operator might direct a drone to reposition itself for a medical package delivery or to conduct a closer inspection of a structure when environmental conditions, mission priorities, or nuanced situational cues cannot be reliably interpreted by the autonomous system alone.

To support these forms of interaction, a significant body of work has explored mechanisms through which humans provide directives to UAVs, especially rotorcraft platforms capable of hovering and maneuvering with high precision in close proximity to people and objects. Several researchers have developed gesture-based systems, where real-time hand poses are mapped to flight commands for single UAVs \cite{yoo, hu3, abdallah} or even swarms \cite{hu2, wijaya}. Others have used voice commands to enable hands-free communication \cite{zhu, pourmehr}, and more recently multimodal fusion of gesture and voice inputs has been shown to outperform either modality alone, with each communication channel compensating for the other’s environmental limitations \cite{aboyie, xiaojia}. However, this prior work has focused primarily on using Computer Vision (CV) or Natural Language Processing (NLP) to recognize human intent and to map it to UAV actions, with the underlying assumption that human-issued commands can be executed safely and directly. This introduces very real safety concerns, as humans may issue incorrect or unclear commands due to limited visibility, loss of spatial awareness, distractions, or reckless behavior, causing the UAV to fly into the terrain, crash into an operator on the ground, or collide with another UAV in the air. The problem is further exacerbated because research experiments are often conducted in controlled or synthetic environments \cite{guyon, hu, aboyie}, where simplified conditions and limited environmental variability may obscure safety risks and overestimate how well these approaches generalize to real-world operational settings.

Despite the significant safety risks associated with direct command execution, existing safety-oriented work has primarily focused on issues such as human comfort, physical separation distances, and safe operating envelopes during human-UAV interaction rather than on the problem of constraining and mediating operator-issued commands prior to execution \cite{zhu}. Addressing this problem requires more than accurate recognition of operator intent; it requires runtime mechanisms capable of determining whether a requested maneuver remains admissible under current environmental conditions, vehicle capabilities, mission objectives, airspace constraints, and team context. This introduces a runtime requirements engineering challenge in which operator inputs must be treated not as unconditional commands, but as high-level intent requests subject to continuous constraint evaluation and safety monitoring.

In this paper, we address these safety concerns by introducing a constrained multimodal operating model in which operator inputs are interpreted as requests for maneuver primitives rather than as direct actuation commands. Each requested maneuver is evaluated against terrain, separation, and flight-envelope constraints prior to execution and continuously monitored during execution. Beyond the operational model itself, we formulate the command mediation process as a requirements-based specification model in which maneuver primitives are associated with explicit preconditions, invariants, guard conditions, and postconditions governing admissibility, execution safety, and emergency handling. These requirements provide a foundation for runtime verification and future reactive synthesis approaches while preserving intuitive human interaction with the autonomous system.

The remainder of the paper is laid out as follows. Section \ref{sec:related} describes related work. Section \ref{sec:model} presents the maneuver-response model, including bounded maneuver primitives, admissibility and safety constraints, the request–evaluate–execute pipeline, and runtime monitoring mechanisms, while Section \ref{sec:reqs} then describes the requirements-governed operationalization of the model through safety properties, runtime guards, state-machine behavior, and illustrative maneuver contracts.
Section \ref{sec:requirements} reports on preliminary experiments and their results. The paper concludes in Sections \ref{sec_threats} and \ref{sec:conclusions} with a discussion of threats to validity, conclusions, and future work.
\section{Related Work}
\label{sec:related}
Related work primarily falls under the topics of multimodal inputs for UAV controls, including speech, gestures, and hardware or software interfaces.

\subsection{Multi-Modal Implementation}
Multimodal approaches combine modalities such as speech and hand gestures to create more flexible UAV control systems. Although gestures are not strictly hands-free, they reduce reliance on handheld controllers or screen-based input, enabling more natural UAV interaction in field settings \cite{zhu, pourmehr}, and both approaches have been shown to outperform traditional joystick interfaces \cite{yoo, villame}. Prior work has demonstrated that combining modalities can improve performance as each performs better under different environmental conditions \cite{aboyie, xiaojia}. For example, speech may remain effective in low-light conditions, whereas gestures can be more reliable in noisy environments \cite{aboyie}. Multi-modal systems also enable cross-modal confirmation, where one modality can help validate another and reduce misinterpretation \cite{xiaojia, aboyie}.
However, prior evaluation has primarily been in controlled indoor or synthetic environments without addressing the variability of real-world operations \cite{aboyie, hu}. As a result, deployment in complex operational settings has been limited, and explicit runtime safety constraint mechanisms have not been rigorously addressed \cite{aboyie, hu, zhou}.

\subsection{Speech Recognition}
Voice recognition offers an intuitive alternative to traditional remote-control and tactile UAV interfaces and can be effectively integrated into UAV control pipelines through automatic speech recognition (ASR) frameworks \cite{contreras}. However, environmental noise and signal distortion remain significant challenges. These issues become even more pronounced in collaborative operational environments, where robust recognition may require techniques such as hidden Markov models (HMMs), deep neural network (DNN) acoustic models, grammar-based syntax analysis, and semantic post-processing to reduce ambiguity in safety-critical settings \cite{park}. More recently, researchers have explored the use of large language models (LLMs) in UAV speech-control pipelines to support more flexible natural-language interaction \cite{choutri}. These systems can translate free-form and spoken instructions into executable UAV commands using hybrid STT and LLM-based architectures \cite{choutri}. Although promising, current approaches still rely heavily on cloud-based inference and remain largely evaluated in controlled or simulated environments, leaving robust real-world deployment as an open challenge \cite{choutri}.

\subsection{CV Gesture Recognition}
CV techniques provide an alternative approach for real-time recognition of human motions and gestures. While early approaches relied on explicit feature extraction, recent work has used deep-learning techniques to provide faster and more robust recognition \cite{starner, cui, chang, abdallah}. Depth-sensing approaches capture three-dimensional positional information \cite{shotton, ren, keskin}, while modern convolutional neural network (CNN) methods learn spatial-temporal features directly from images and video streams \cite{molchanov, kopuklu, liu}. However, these systems remain sensitive to environmental factors such as lighting and occlusion \cite{wachs}, and models trained on controlled datasets often fail to generalize to real-world environments \cite{guyon, wang}.
Gesture-recognition systems have been applied to UAV control by mapping hand poses to commands such as takeoff, landing, and directional movement \cite{yaseen}. These systems often rely on small gesture vocabularies and controlled datasets \cite{yaseen}. Gesture control has also been extended to UAV swarms, where a single operator may coordinate multiple drones simultaneously \cite{hu2, wijaya}. However, scalability and safety remain major challenges, since a single misinterpreted command may affect multiple UAVs at once \cite{hu2, wijaya}. While frameworks such as ROS2 support scalable implementations \cite{kondratev}, current approaches still provide limited support for uncertainty propagation and runtime safety constraint evaluation \cite{wang2}. Consequently, system reliability depends not only on recognition accuracy, but also on how failures and ambiguities are handled at the control level \cite{obaid, cauchard}.

\subsection{The Need for Runtime Safety Mediation}
Taken together, these multimodal approaches demonstrate significant promise for enabling intuitive and flexible UAV interactions. However, speech, gesture, and GUI inputs remain inherently uncertain and susceptible to misinterpretation due to environmental conditions, ambiguity, sensor limitations, and human error. Consequently, reliable human-UAV interaction depends not only on accurate intent recognition, but also on runtime mechanisms capable of evaluating whether requested maneuvers remain safe and admissible within the current operational context \cite{benford2025somaticsafetyembodiedapproach, DBLP:conf/seams/Cleland-HuangAV22, hri-safety1, hri-safety2}.

\section{Multi-Modal Maneuver Request Model}
\label{sec:model}

\begin{table*}[b]
\centering
\caption{Baseline maneuver requests are mapped to bounded intentions. The multimodal requests can be provided by gestures, verbal commands, or through a GUI.}
\label{tab:nudger-primitives}
\setlength{\tabcolsep}{6pt}
\begin{tabular}{|l|l|l|l|}
\hline
\textbf{Maneuver Request} & \textbf{Primitive Action} & \textbf{Nominal Intent} & \textbf{Example Bounds} \\
\hline
Go left & MoveLeft & Lateral translation left &
$\leq 1\,\mathrm{m} \land (velocity < x\,\mathrm{m/s})$ \\
Go right & MoveRight & Lateral translation right &
$\leq 1\,\mathrm{m} \land (velocity < x\,\mathrm{m/s})$ \\
Go forward & MoveForward & Forward translation &
$\leq 1\,\mathrm{m} \land (velocity < x\,\mathrm{m/s})$ \\
Go back & MoveBackward & Rearward translation &
$\leq 1\,\mathrm{m} \land (velocity < x\,\mathrm{m/s})$ \\
Go up & Ascend & Vertical translation upward &
$\leq 0.5\,\mathrm{m} \land (velocity < x\,\mathrm{m/s})$ \\
Go down & Descend & Vertical translation downward &
$\leq 0.5\,\mathrm{m} \land (velocity < 0.5x\,\mathrm{m/s})$ \\
Turn left/right & ControlledYawScan & Slow full rotation, return &
$= 360^\circ \land (\dot{\psi} < \omega_{\mathrm{scan}}\,\mathrm{deg/s}) \land \psi_{\mathrm{final}}=\psi_A$ \\
&&to operator-facing orientation&\\
\hline
\end{tabular}
\end{table*}

In this section we present our Multi-Model Maneuver Request (M3R) model. 
At a high-level, the UAV is always in one of the following modes. 
\begin{itemize}
    \item \textbf{NormalAutonomy}: The drone performs its mission under nominal autonomous control. This high-level mode encompasses a variety of lower level states such as takeoff, land, fly-to-waypoints, and circle. In this mode the drone is     \textbf{disengaged} from direct interaction with the human operator. 
    \item\textbf{ManeuverResponseArmed}: The drone is eligible to receive maneuver commands, but no command is currently being acted upon.
    \item \textbf{ManeuverResponseActive}: A maneuver request has been accepted and translated into a bounded motion primitive.
    \item \textbf{Blocked}: A requested maneuver has been rejected because it would violate one or more safety constraints.
    \item \textbf{EmergencyHold}: The UAV has interrupted execution of an MRM maneuver and transitioned to a safe hover or hold state.
    \item \textbf{ReturnToAnchor}: The drone returns to a safe anchor point after a maneuver or interruption.
\end{itemize}

\subsection{Runtime Pairing of Operator and UAV}
Pairing occurs between a single operator and a single UAV when the operator sends a {\it pairing request} over existing communication channels (e.g., telemetry or mesh-radio). The request contains the operator’s coordinates (latitude, longitude, and altitude above mean sea level (AMSL)). After receiving the request, the UAV rotates to face the operator, transitions into the \textit{ManeuverResponseArmed} state, and returns a confirmation message that may include visual or auditory cues in addition to standard communication mechanisms.
As the UAV is required to maintain an operator-facing orientation, all subsequent directional commands are interpreted relative to the operator’s perspective rather than the UAV body frame. For example, if the operator instructs the UAV to ``move right,'' the UAV moves to its left. This design choice reduces ambiguity during real-world operations, particularly at longer distances where operators may struggle to accurately perceive the UAV’s heading and orientation.

\subsection{Mapping Maneuver Commands to Primitive Actions}
The M3R model allows the UAV to act upon a finite set of  requests, each of which is mapped to a bounded motion primitive. The current set of mappings are depicted in Table~\ref{tab:nudger-primitives}.

Each primitive includes the following attributes:
\begin{itemize}
    \item intended direction,
    \item maximum displacement,
    \item maximum speed,
    \item maximum duration,
    \item completion condition,
    \item safety preconditions, and
    \item abort conditions.
\end{itemize}

\subsection{Request--Evaluate--Execute Pipeline}
Each maneuver request passes through the following stages:
\begin{enumerate}
    \item \textbf{Interpret}: Process the input request (e.g., gesture, speech, or GUI interaction), compute confidence in the interpretation of the request, and if confidence is sufficient, infer the intended maneuver. We establish a {\it confidence-threshold} ($cf$) for all examples and experiments reported in this paper.
    \item \textbf{Validate}: Check whether the requested maneuver is allowed in the current context. For example, if the operator requests the UAV to {\it move-right}, but there is a steep incline in that direction, the UAV will tag the request as {\it unsafe}.
    \item \textbf{Constrain}: In the case of an unsafe maneuver, the UAV either prohibits the maneuver or reduces it to a safe bounded version.
    \item \textbf{Execute}: Perform the approved motion primitive.
    \item \textbf{Monitor}: Continuously evaluate safety constraints during execution.
    \item \textbf{Complete or Abort}: Terminate the maneuver when complete or interrupt it if a violation is detected.
\end{enumerate}

\subsection{Anchor-Based Local Workspace}
When the {\it ManeuverResponseArmed} mode is initially activated, the UAV records its current position and attitude as its {\it anchor state}:
\[
A = (x_A, y_A, z_A, \psi_A)
\]
where $x_A$, $y_A$, and $z_A$ transform latitude, longitude, and altitude into a local coordinate systems, and $\psi_A$ represents the direction the UAV is facing (heading). Notably this must be towards the operator with which the UAV is paired. Each subsequent maneuver is constrained to remain within a bounded local workspace around this anchor, defined as
\[
(x - x_A)^2 + (y - y_A)^2 + (z - z_A)^2 \leq R_{\max}^2
\]
where, for example $R_{\max} = 5\,\mathrm{m}$.

This workspace may be further clipped by adjacent terrain, nearby drones, or geofencing constraints. Each time a new command is received by the UAV, it dynamically updates its anchor and subsequently the boundaries of its allowed workspace. 

\subsection{Safety Constraint Categories}
In addition to providing assurances that the UAV remains within its bounded local workspace, the safety model evaluates each requested maneuver against four complementary categories of constraints derived from common UAV failure modes and operational risks (e.g., \cite{DBLP:conf/sigsoft/VierhauserIACM21}).\vspace{4pt}

\noindent{$\bullet$ \bf Terrain Constraints:}
Terrain constraints ensure that the UAV maintains safe clearance relative to the ground and local topography. Example conditions include:
\begin{align*}
\mathrm{AGL}(s') &\geq \mathrm{MinAGL} \\
\mathrm{TerrainClearance}(\pi) &\geq C_{\min}
\end{align*}
where $s'$ is the predicted future state and $\pi$ is the predicted path for the requested maneuver. The first constraint ensures that the UAV does not descend below a minimum allowable altitude above ground level, while the second ensures that the predicted maneuver path maintains sufficient clearance from terrain and obstacles. 
\vspace{4pt}

\noindent{$\bullet$ \bf Inter-UAV Separation Constraints:}
Inter-drone constraints ensure safe spacing and prevent conflicts with nearby drones. For each neighboring drone $d \in D$:
\[
\mathrm{Sep3D}(\pi_{\selfm}, \pi_d) \geq S_{\min}
\]
where $\pi_{\selfm}$ is the predicted path of the requesting drone and $\pi_d$ is the predicted path or reserved path volume of drone $d$. This constraint ensures that the minimum three-dimensional separation between the two predicted trajectories remains above a required safety threshold throughout maneuver execution.  
\vspace{4pt}

\noindent{$\bullet$ \bf Flight Envelope Constraints:} 
Flight envelope constraints bound the magnitude, velocity, and rotational rate of each maneuver:
\begin{align*}
|\Delta x| &\leq X_{step} \\
|\Delta y| &\leq Y_{step} \\
|\Delta z| &\leq Z_{step} \\
 v &\leq V_{\max}^{nudger} \\
 \dot{\psi} &\leq \Omega_{\max}^{nudger}
\end{align*}
where $\Delta x$, $\Delta y$, and $\Delta z$ represent the requested positional displacement along each axis, $v$ denotes translational velocity, and $\dot{\psi}$ denotes yaw rate. These constraints ensure that MR interactions remain bounded supervisory maneuvers rather than unconstrained teleoperation.
\vspace{4pt}

\noindent{$\bullet$ \bf Context and Confidence Constraints:}
Context and confidence constraints determine whether the UAV transitions from {\it ManeuverResponseArmed} to {\it ManeuverResponseActive} mode under current operating conditions:
\begin{align*}
\mathrm{InputConfidence}(m) &\geq C_{\min}^{(m)} \\
\mathrm{LocalizationConfidence} &\geq L_{\min} \\
\mathrm{TerrainModelConfidence} &\geq T_{\min} \\
\mathrm{NeighborStateAge} &\leq \tau_{\max}
\end{align*}
where $m$ denotes the input modality (e.g., gesture, speech, or GUI), $\mathrm{InputConfidence}(m)$ represents the confidence associated with the interpreted request from that modality, and $\mathrm{NeighborStateAge}$ denotes the elapsed time since the most recent state update for nearby agents. These constraints ensure that maneuvers are only permitted when the system possesses sufficiently reliable input interpretation, localization accuracy, terrain awareness, and neighboring drone state information. A maneuver is blocked if any condition is violated.

\begin{figure}
    \centering
    \includegraphics[width=\linewidth]{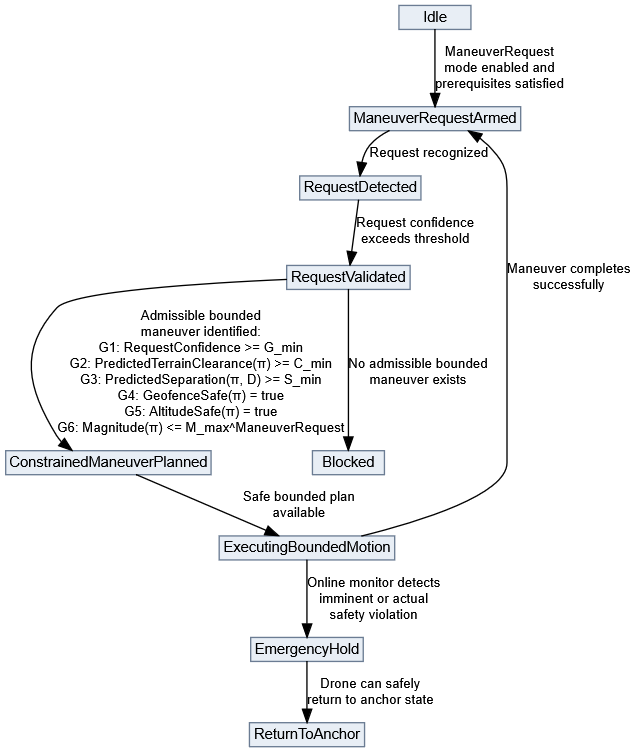}
    \caption{State machine operationalizing the M3R model, including request interpretation, safety validation, constrained maneuver planning, execution, and runtime intervention.}
    \label{fig:MR-statemachine}
\end{figure}
\subsection{Safety Decision Function}
When a UAV receives a maneuver request, it analyzes the request to determine whether the maneuver can be executed safely within the current operational context. The UAV performs this analysis using the following safety-decision function. 
Let
\begin{itemize}
    \item $g$ denote the recognized MR,
    \item $x$ denote the current UAV state,
    \item $T$ denote the terrain model,
    \item $D$ denote the set of neighboring UAVs, and
    \item $C$ denote contextual confidence values.
\end{itemize}
The system computes:
\begin{align*}
\mathrm{intent} &= \mathrm{Interpret}(g) \\
\pi_{cand} &= \mathrm{Plan}(\mathrm{intent}, x) \\
\pi_{safe} &= \mathrm{Filter}(\pi_{cand}, T, D, C)
\end{align*}
A candidate maneuver is admissible only if
{\raggedright
\[
\begin{aligned}
\mathrm{Safe}(\pi_{cand}, x, T, D, C) :=
&\ \mathrm{TerrainSafe} \wedge \\ \mathrm{SeparationSafe} &\ \wedge \\
\mathrm{EnvelopeSafe} \wedge \\
\mathrm{ConfidenceAdequate}.
\end{aligned}
\]
}
If $\mathrm{Safe}(\pi_{cand}, x, T, D, C)$ is false, the maneuver is either reduced, delayed, rejected, or replaced with a safer alternative.

\section{Requirements-Guided Operationalization}
\label{sec:reqs}

The maneuver-response model presented in the previous section provides a conceptual framework for mediating human intent into safe UAV motion. In this section, we show how the model can be translated into implementation-oriented artifacts through a requirements-guided operationalization process. We first derive a formal requirements specification that captures the behavioral and safety properties of the maneuver-response model. We then present two complementary operationalizations of this specification: a state-machine representation describing system-level behavior and maneuver-specific contracts defining the requirements associated with individual maneuver primitives. Together, these artifacts provide a traceable path from high-level safety requirements to executable controller behavior while supporting verification and future automated controller synthesis.

\subsection{Requirements Specification}

The maneuver-response model is first expressed as a set of behavioral and safety requirements. These requirements are derived from the preconditions, invariants, guard conditions, and postconditions associated with each maneuver primitive, and constrain the UAV state, terrain model, operator input confidence, and neighboring UAVs. Collectively, they define the expected behavior of the maneuver-response system independently of any implementation.

At a high level, the specification enforces the following properties:

\begin{itemize}[leftmargin=*]
    \item {\it ManeuverResponseArmed} mode is activated only after successful pairing has occurred between the operator and the UAV.
    \item A maneuver request is validated only when the associated input confidence exceeds the required threshold and the candidate maneuver satisfies all applicable safety constraints.
    \item When an intended maneuver fails to satisfy spatial or environmental safety constraints, the requested maneuver is iteratively shortened along its original intended path until a safe bounded trajectory is identified. The maneuver is rejected if no admissible trajectory exists.
\end{itemize}

For example, the second requirements above (i.e., low-confidence requests must never activate maneuver execution) can be expressed formally in signal temporal logic \cite{} as:
\[
\begin{aligned}
\Box \Big(
&(\mathrm{ManeuverRequestDetected} \\
&\wedge\ \mathrm{GestureConfidence} < G_{\min}) \\
&\rightarrow \bigcirc(\texttt{Blocked})
\Big)
\end{aligned}
\]
where $\Box$ denotes the ``always'' operator and $\bigcirc$ denotes the ``next-state'' operator. 

The resulting requirements provide a formal specification against which operational models and implementations can be evaluated. Requirements governing maneuver admissibility, safety constraints, workspace bounds, and emergency handling support systematic analysis, verification, testing, and traceability throughout the development process. They also establish a foundation for future requirements-driven controller synthesis, in which environmental assumptions and system guarantees may be used to automatically construct correct-by-design maneuver-response controllers. In the following subsection, we present a baseline state-machine realization that operationalizes this specification.

\subsection{State Machine Representation}

One operational realization of the requirements specification is a maneuver-response state machine. The state machine provides an executable behavioral model in which each transition corresponds to the satisfaction of one or more requirements defined in the previous subsection. Such a representation may be developed manually as part of the software engineering process or generated automatically through future controller synthesis techniques. A baseline realization is presented below and illustrated in Figure~\ref{fig:MR-statemachine}.

{\raggedright
\begin{itemize}
    \item $\texttt{Idle} \rightarrow \texttt{ManeuverRequestArmed}$ when MR mode is enabled and prerequisites are satisfied.
    \item $\texttt{ManeuverRequestArmed} \rightarrow \texttt{ManeuverRequestDetected}$ when a gesture is recognized.
    \item $\texttt{ManeuverRequestDetected} \rightarrow \texttt{ManeuverRequestValidated}$ when gesture confidence exceeds threshold.
    \item $\texttt{ManeuverRequestValidated} \rightarrow \texttt{ManeuverPlanned}$ when safety guards are satisfied.
    \item $\texttt{ManeuverRequestValidated} \rightarrow \texttt{Blocked}$ when one or more safety constraints fail.
    \item $\texttt{ManeuverPlanned} \rightarrow \texttt{ExecutingBoundedMotion}$ when a safe plan is available.
    \item $\texttt{ExecutingBoundedMotion} \rightarrow \texttt{EmergencyHold}$ when an online monitor detects imminent or actual safety violation.
    \item $\texttt{ExecutingBoundedMotion} \rightarrow \texttt{NudgerArmed}$ when the maneuver completes successfully.
    \item $\texttt{EmergencyHold} \rightarrow \texttt{ReturnToAnchor}$ when the drone can safely return to the anchor state.
\end{itemize}
}

Representative guard conditions for transition to \texttt{ManeuverPlanned} are:
\begin{align*}
G_1 &: \mathrm{GestureConfidence} \geq G_{\min} \\
G_2 &: \mathrm{PredictedTerrainClearance}(\pi) \geq C_{\min} \\
G_3 &: \mathrm{PredictedSeparation}(\pi, D) \geq S_{\min} \\
G_4 &: \mathrm{GeofenceSafe}(\pi) = \mathrm{true} \\
G_5 &: \mathrm{AltitudeSafe}(\pi) = \mathrm{true} \\
G_6 &: \mathrm{Magnitude}(\pi) \leq M_{\max}^{nudger}
\end{align*}

These guard conditions correspond to runtime-monitorable signals and operationalize the behavioral and safety requirements defined in the previous subsection. Consequently, the state machine provides a concrete implementation model whose execution can be verified against the underlying requirements specification.

\subsection{Contract-Based Operationalization}

A complementary operationalization of the requirements specification is provided through maneuver-specific behavioral contracts. Rather than describing system-wide behavior as a state machine, contracts specify the obligations associated with individual maneuver primitives in terms of preconditions, invariants, and postconditions. The following example illustrates the contract associated with the \texttt{GoForward} maneuver.

\vspace{6pt}

\noindent\textsc{\textbf{GoForward}}

\noindent\textbf{Preconditions}
\begin{itemize}
    \item[-] nudger mode is active,
    \item[-] gesture confidence is above threshold,
    \item[-] localization is healthy,
    \item[-] terrain and neighbor state data are sufficiently fresh.
\end{itemize}

\textbf{Invariants}
\begin{itemize}
    \item[-] terrain clearance remains above minimum threshold,
    \item[-] inter-drone separation remains above minimum threshold,
    \item[-] motion remains inside the bounded nudger workspace,
    \item[-] speed remains below the nudger-mode speed bound.
\end{itemize}

\textbf{Postconditions}
\begin{itemize}
    \item[-] the drone reaches an approved final state within the local workspace, or
    \item[-] the maneuver is safely aborted to hover or hold.
\end{itemize}

These contracts provide a direct realization of the requirements specification at the maneuver level. Preconditions define when execution may begin, invariants capture the safety properties that must be preserved throughout execution, and postconditions characterize acceptable completion or safe termination. Together with the state-machine representation, they provide complementary operational models that can be verified against the common requirements specification while maintaining traceability from high-level safety requirements to implementation.

\section{Implementing the Model: An Initial Prototype}
\label{sec:requirements}
While our model lends itself to both formal verification and reactive control synthesis, for purposes of this workshop paper we have implemented it using a manual requirements-driven design and validation process.

This separation ensures that safety does not depend exclusively on perfect human input recognition.

\begin{figure*}[t]
    \centering
    \includegraphics[width=\textwidth]{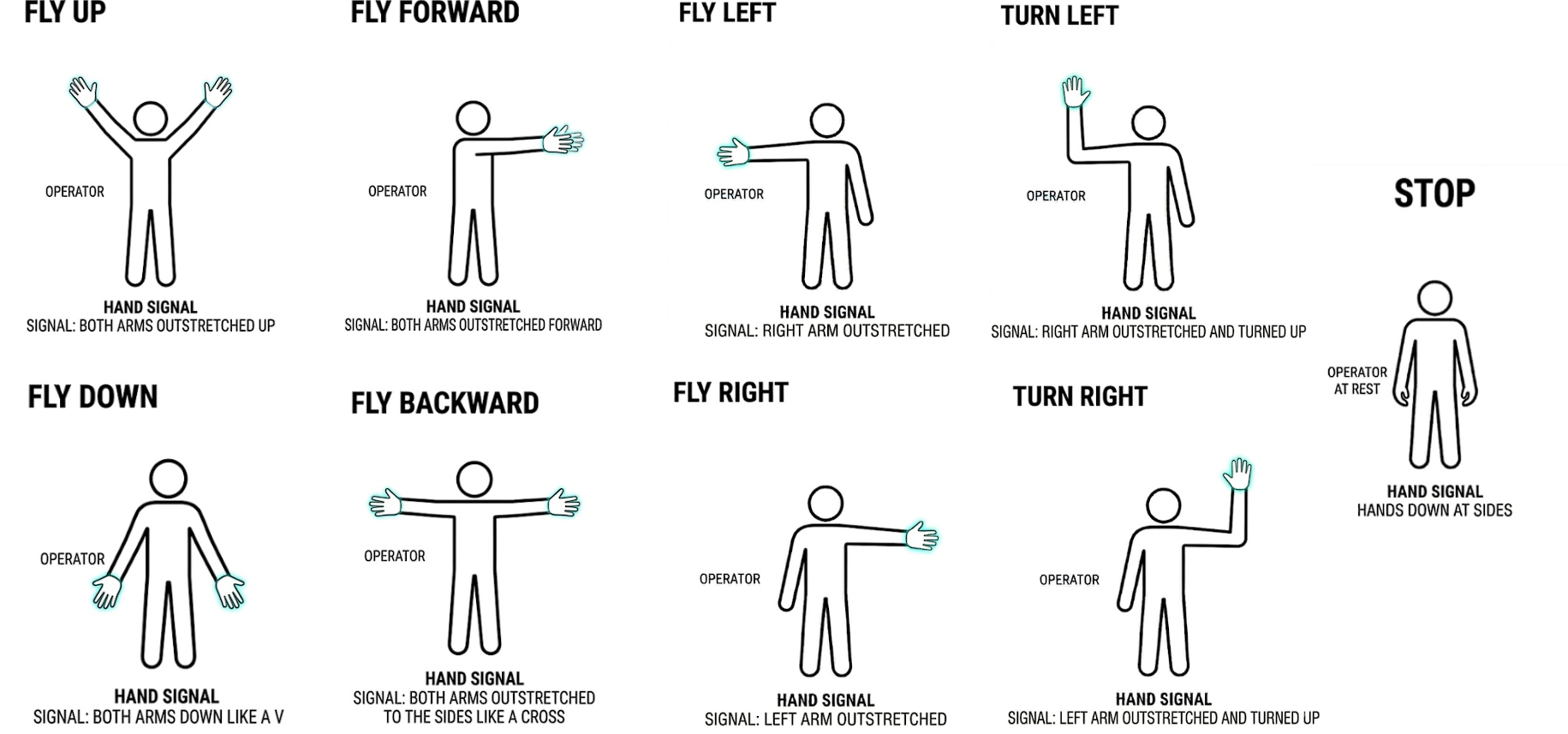}
     \caption{Motion primitives for gesture-based commands. Each command supports ``Fly'' or ``Go'' prefix, but are shown here as `Fly' commands.}
    \label{fig:gestures}
\end{figure*}

\subsection{Requirements}
We translated the model into a baseline set of requirements which which were then used to design, implement, and validate the DR-Maneuvers for our input modalities.  The requirements are specified as follows using Easy Requirements Specification (EARS) notation \cite{EARS}.

\begin{enumerate}
    \item When an MR-Maneuver input is recognized, the Maneuver-Request Interpretation Layer shall interpret the input as a request for a bounded motion primitive.

    \item When a maneuver request is interpreted, the MR-Planner shall generate a bounded candidate maneuver plan.

    \item When a candidate maneuver plan is generated, the Safety Supervisor shall evaluate the plan against predicted terrain clearance before execution.

    \item When execution of a maneuver plan would reduce terrain clearance below the configured minimum safe threshold, the Safety Supervisor shall reject or constrain the maneuver plan.

    \item When a candidate maneuver plan is generated, the Safety Supervisor shall evaluate the plan against the predicted positions or reserved path volumes of nearby UAVs.

    \item When a maneuver plan would violate minimum inter-UAV separation constraints, the Safety Supervisor shall reject or constrain the maneuver plan.

    \item While ManeuverResponse mode is active, the MR-Planner shall constrain maneuver plans to remain within the bounded local workspace relative to the current anchor state.

    \item While ManeuverResponse mode is active, the MR-Planner shall limit maneuver displacement, velocity, and duration to configured flight-envelope bounds.

    \item While a maneuver is executing, the Runtime Monitor shall continuously monitor terrain clearance, inter-UAV separation, and workspace-bound violations.

    \item When a hard safety constraint is violated or predicted to be violated during maneuver execution, the Runtime Monitor shall command the Motion Controller to abort the maneuver and transition the UAV into the EmergencyHold state.

    \item When localization confidence, input confidence, terrain-model confidence, or neighboring-UAV state freshness falls below required thresholds, the Safety Supervisor shall reject or constrain the requested maneuver.
\end{enumerate}

\subsection{Architecture}
The prototype architecture separates out input detection and interpretation from safety enforcement. A baseline decomposition includes:
\begin{itemize}[leftmargin=*]
    \item \textit{Maneuver-Request  Interpretation Layer}: recognizes gestures, speech, and GUI inputs, and infers requested maneuvers,
    \item \textit{MR-Planner}: converts intentions into local bounded candidate plans,
    \item \textit{Safety Supervisor}: evaluates plans against terrain, drone separation, geofencing, and confidence constraints,
    \item \textit{Motion Controller}: executes only approved maneuvers, and
    \item \textit{Runtime Monitor}: continuously checks safety invariants during execution.
\end{itemize}

\subsection{Modality Interfaces}
We have designed initial interfaces for all three modalities and implemented prototypes that have undergone preliminary tests in a lab environment with a simple simulated drone. 

\subsubsection{Gesture-Based Maneuvers}
Our current design supports nine unique commands as described in Table \ref{tab:nudger-primitives}. We recognize all nine gestures in the current model but limit execution to Fly/Go Up, Down, Forward, Left, Right,  Backward, Forward, and Stop. While Turn Left and Turn Right are conceptually simple to enact, they require additional work (outside the scope of this workshop paper) to assure that (a) when operating purely in gesture mode the drone always returns to face the operator, or (b) when operating in multimodal input mode (e.g., Gesture + Voice) commands such as `Go Left' are enacted relative to the operator regardless of direction the drone is facing during the turn. For purposes of the initial prototype we used the Visual Large Language Model (VLMM) capabilities of GPT 5.o to recognize each of the human gestures in a video stream. This will be replaced in later work with a 
Yolo-trained model to recognize gestures at distance and altitude in potentially inclement conditions using our NOMAD data set \cite{nomad}. 
\begin{figure}[h]
    \centering
    
    \begin{subfigure}{0.3\columnwidth}
        \centering
        \includegraphics[width=\linewidth]{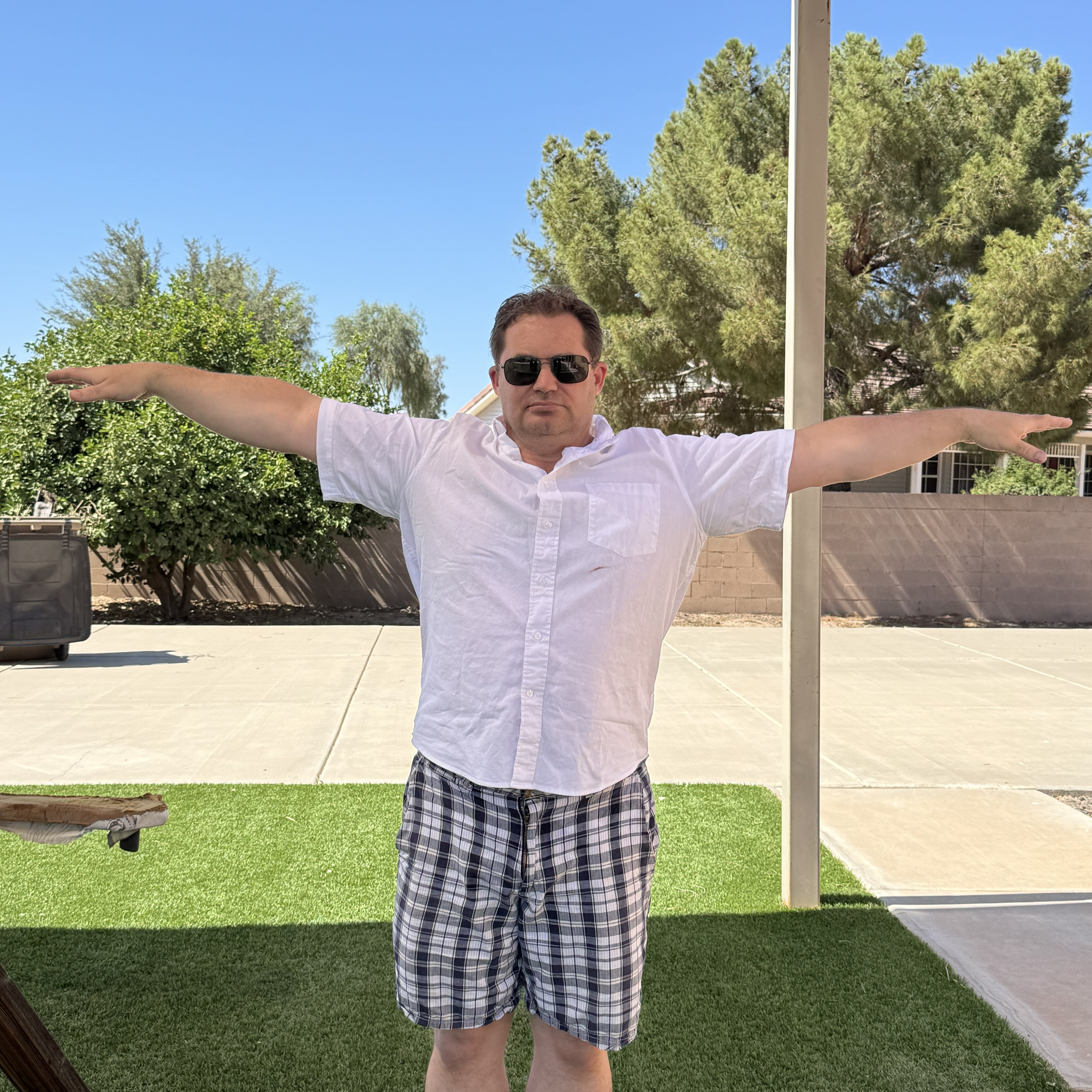}
        \caption{Go Back}
    \end{subfigure}
    \hfill
    \begin{subfigure}{0.3\columnwidth}
        \centering
        \includegraphics[width=\linewidth]{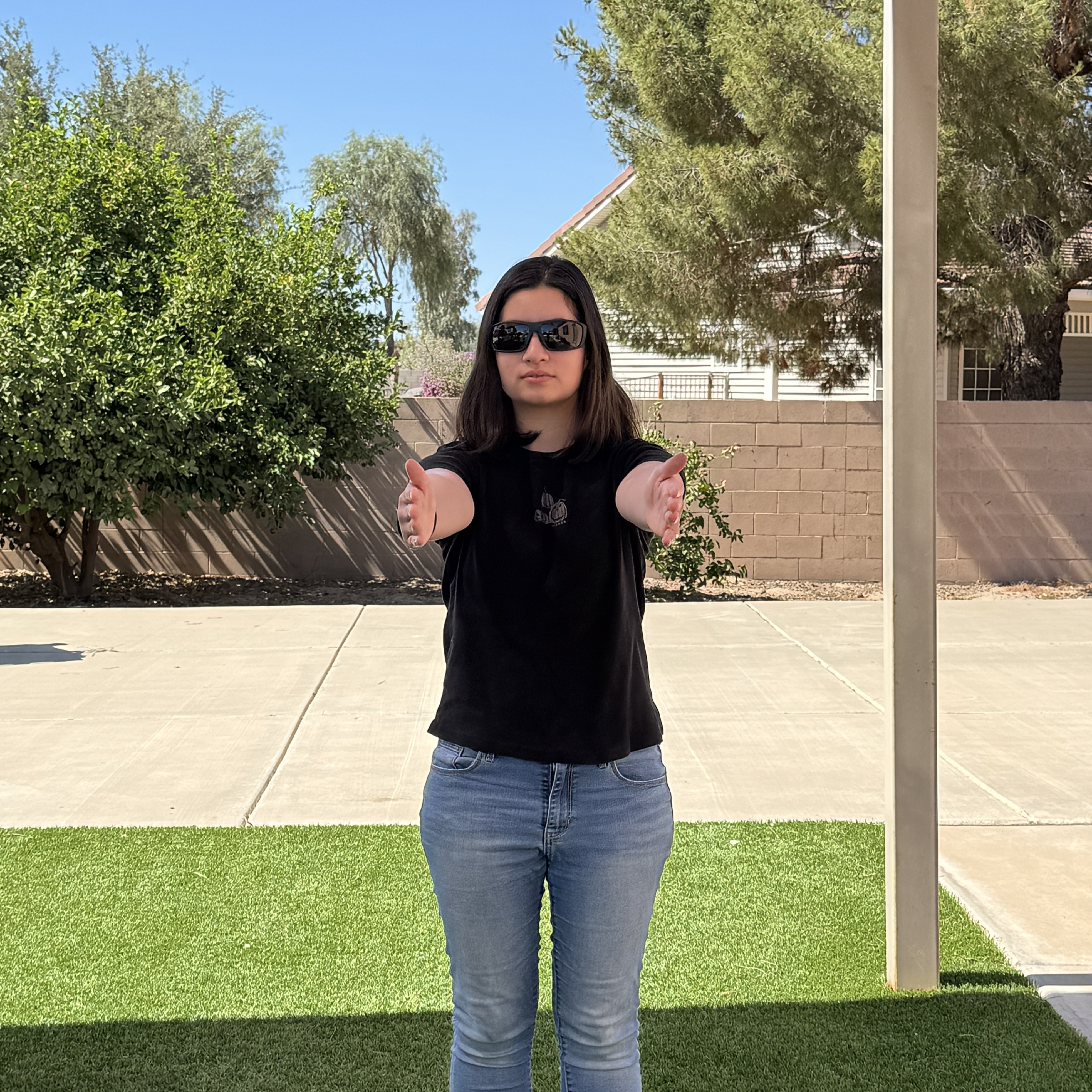}
        \caption{Go Forward}
    \end{subfigure}
    \hfill
    \begin{subfigure}{0.3\columnwidth}
        \centering
        \includegraphics[width=\linewidth]{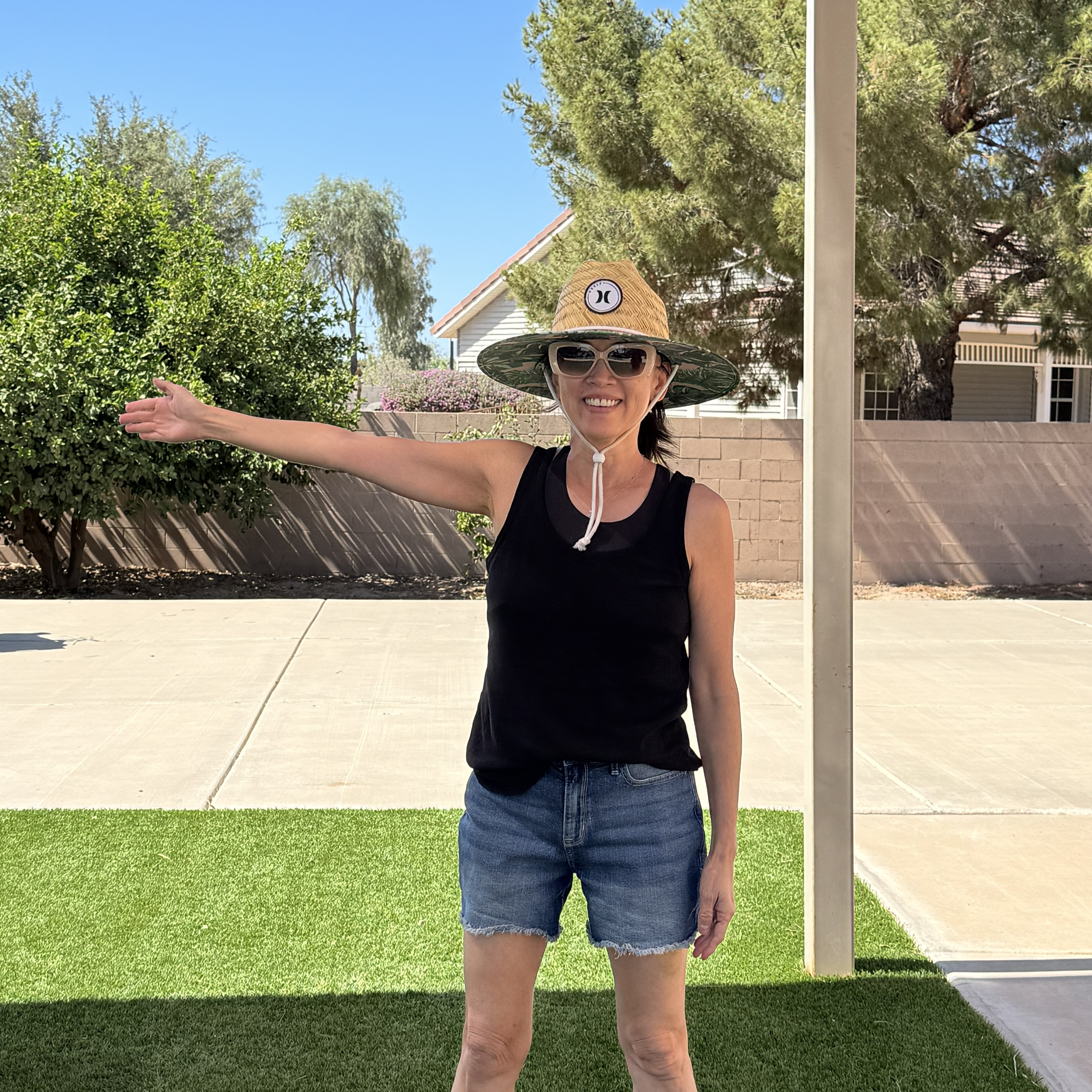}
        \caption{Go Right}
    \end{subfigure}

    \caption{A sequence of motion commands issued by human operators in an outdoor setting, and tested for recognition by the VLLM. All commands are from the user's perspective, therefore Go Right is executed by the drone moving to its left as it faces the human.}
    \label{fig:motion-sequence}
\end{figure}

\subsubsection{Voice-Based Maneuvers}
Voice activated maneuvers are recognized using VOSK, an open-source, offline speech recognition toolkit built on the Kaldi speech recognition framework. VOSK uses acoustic models trained with Kaldi’s toolchain, which is based on deep neural networks for acoustic modeling combined with n-gram language models for decoding. The chosen model is \texttt{vosk-model-en-us-0.22}, a mid-sized (1.8GB) model trained on American English acoustic data. 

The audio is collected from a microphone, and is captured continuously at 16 kHz, mono, 16-bit PCM in chunks of 4000 frames ($\sim$250ms each), which are processed individually using VOSK’s \texttt{KaldiRecognizer}. The recognizer is initialized upfront with a JSON grammar list comprised of every ``legal'' phrase, constraining the acoustic model to only consider known commands as candidates and labeling all non-relevant tokens as \texttt{[unk]}. The recognizer collects \textit{partial results} as audio arrives, and then once silence is detected, it outputs \textit{final results} matching a term in the grammar. NLP cleaning is applied to filter out noise, and then three sequential pattern checks are run in priority order:

\begin{enumerate}
\item Check for activation phrases (“ACTIVATE {color}” or “DISCONNECT {color}” substrings)
\item Check for “STOP/CONTINUE” commands
\item Check for directional phrases, which map every natural variant (“FLY BACKWARDS, FLY BACK, GO BACKWARD”, etc.) to a single canonical direction (e.g., “\texttt{back}"). The first match “wins,” and is returned. If nothing matches, \texttt{None} is returned, and the utterance is discarded.
\end{enumerate}

Given this implementation, \textit{Voice Activation Maneuvers} are only generated in cases when confidence is high that they match a targeted key-term. 

\subsection{Graphical User Interface Inputs}
Our third modality involves a simple User Interface with buttons representing each of the actions shown in Figure \ref{fig:gestures}. In future implementations we will add addition hardware interfaces such as a joystick.  

\subsection{Multi-Model Inputs}
\begin{figure}
    \centering
    \includegraphics[width=\linewidth]{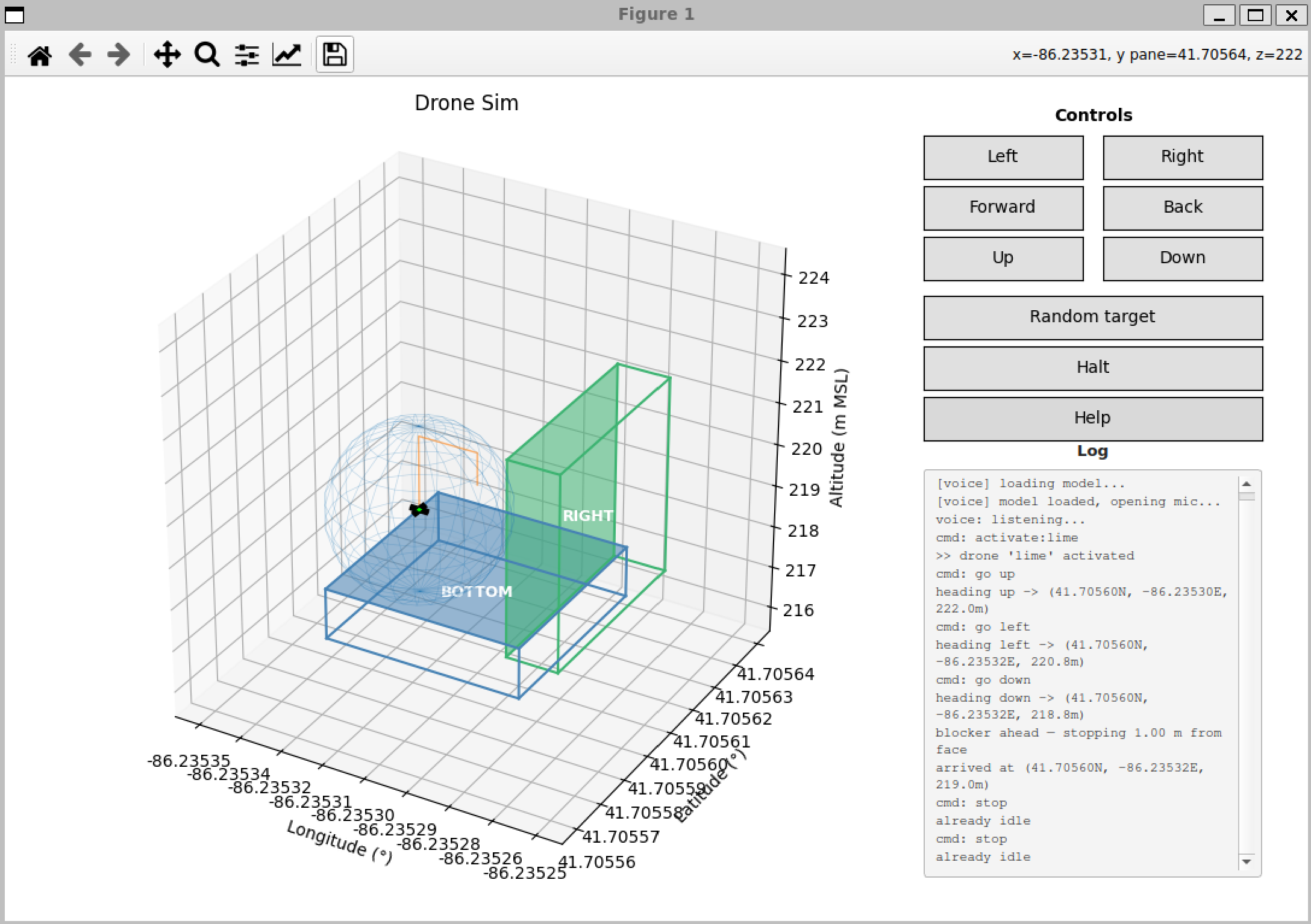}
    \caption{Our prototype tool implements Speech, Gestures, and GUI-based commands interchangeably.  The model updates its anchor position upon completion or interruption of each command, and halts at a predefined distance from terrain barriers (shown here in blue and green).}
    \label{fig:tool}
\end{figure}

Finally, we built a prototype tool which accepts inputs from any of our three modalities interchangeably.  The tool is shown in Figure \ref{fig:tool}.  The tool was intended for initial validation purposes only. It was built using Python and MatPlotLib and provide a 3D environment defined by a moving window of latitude, longitude, and altitude ranges. The drone itself was represented by a 2D image situated with a sphere representing the workspace. Prior to receipt of a command, the drone was anchored at the center point of the sphere, and the subsequent command was executed relative to that point. 

The operating constraints related to avoiding terrain are supported by an Environmental Digital Twin \cite{DBLP:conf/models/BernalPGMMC25} and avoiding other UAVs by an air-leasing technology that guarantees that under nominal conditions all UAVs operate in unique airspace \cite{chambers2025automated}.

\section{Threats to Validity}
\label{sec_threats}
There are several threats to validity to this work. First and foremost, we have presented the maneuver-response model as a foundation for both requirements-driven development and control synthesis; however, only the requirements-driven path has been translated into a functioning prototype. Our claims regarding control synthesis and formal verification are therefore grounded in the analyzability of the model rather than on a synthesized controller.  Second, the framework reflects a bounded set of maneuver primitives and safety constraints; however, several aspects of the model require additional exploration. For example, fully addressing turn-based maneuvers, was out of scope of this work, as it introduces additional challenges related to maintaining operator-UAV communication during and after rotation, and adds additional implementation-level requirements for addressing new commands that might be issued during a turn. 

A third threat is that for initial prototyping purposes, we opted to use a general-purpose Visual Large Language Model (VLLM) rather than a purpose-built CV model and tested gesture recognition under close-range, ideal conditions leaving gesture recognition performance under field conditions unvalidated. Moving forward, we will conduct a large-scale field-based data collection, and use the collected data to train a CV model with empirically defined capabilities and limitations for accurately recognizing gestures at varying distances and pitches \cite{nomad}.

Fourth, in this paper our focus was on constructing the basic MR-Maneuver model, rather than on rigorous real-world implementation and validation. The model therefore does not yet account for real-world uncertainties such as wind, GPS drift, communication latency, rapidly changing airspace, or degraded visual and acoustic conditions \cite{DBLP:conf/re/GranadenoBIC24}. In future work, we will extend the model to address these real-world phenomenon, and integrate both the synthesized controller and requirements-derived solutions with our existing terrain model \cite{DBLP:conf/models/BernalPGMMC25}) and air-space coordination system \cite{chambers2025automated}.  

Finally, a fundamental claim of our work is that, unlike prior gesture- and speech-based interfaces for UAV maneuvers, we explicitly address runtime safety concerns. However, our current validation remains limited, as it currently only partially addresses real-world phenomena such as geolocation uncertainties, and rigorous field-based evaluation involving both static obstacles (e.g., terrain and structures) and dynamic obstacles (e.g., other UAVs) is needed. At present, our validation has been limited to controlled prototype experiments where obstacles were dynamically introduced into a simulated environment, and the system demonstrated its ability to halt or constrain a human-initiated maneuver when a safety violation is detected. Despite these limitations, the current work establishes the core maneuver-response abstraction, safety mediation pipeline, and requirements framework needed to systematically reason about multimodal human intent prior to execution. We therefore view this work not as a completed operational solution, but as a foundational step toward runtime-assured and formally analyzable human–UAV interaction in complex operational environments.

\section{Conclusions}
\label{sec:conclusions}
Taken together, the threats and limitations discussed in the previous section reinforce a central theme of this work that safe multimodal UAV interaction cannot depend solely on accurate intent recognition. Instead, speech, gestures, and GUI interactions must be mediated through explicit runtime safety constraints prior to execution. To address this problem, we introduced a maneuver-response framework in which operator inputs are interpreted as bounded maneuver requests subject to continuous evaluation against terrain, separation, localization, confidence, and flight-envelope constraints.

Beyond the operational model itself, this work establishes a requirements-governed foundation for runtime assurance and future controller synthesis for human–UAV interaction in complex operational environments. The framework formalizes maneuvers through requirements, safety contracts, runtime guards, and state-machine behavior, thereby supporting analyzability, verification, and future reactive synthesis approaches.

Future work will focus on addressing the primary threats and open challenges identified in this paper, particularly the problem of preserving runtime assurance and analyzability under increasingly realistic operational conditions. This includes integrating terrain-aware reasoning capabilities \cite{DBLP:conf/models/BernalPGMMC25}, scene-understanding pipelines, dynamic airspace-awareness mechanisms, ADS-B data, and DroneResponse’s Air-Leaser framework for fine-grained airspace deconfliction \cite{chambers2025automated} into the maneuver-response pipeline. While many of these operational capabilities have already been developed independently, combining them within a unified maneuver-response framework introduces significant challenges for controller synthesis, runtime monitoring, and safety assurance due to uncertainty in sensing, localization, communication, and environmental dynamics. Evaluation will therefore progress incrementally from controlled studies within our MOMUX environment \cite{olesk2026momux} toward increasingly realistic outdoor field experiments involving physical UAVs operating under real-world terrain, airspace, sensing, and multi-UAV coordination constraints.

% REFERENCES
\IEEEtriggeratref{33}
\bibliographystyle{IEEEtran}
\bibliography{bibby}

\end{document}